\documentclass[preprint]{elsarticle}

\usepackage{float}
\usepackage{geometry}
\usepackage{amsmath,amssymb,amsfonts}
\usepackage{algorithmic}
\usepackage{graphicx}
\usepackage{textcomp}
\usepackage{subfigure}
\usepackage{makecell}
\usepackage{hyperref}
\hypersetup{
    colorlinks=true,
    linkcolor=blue,
    filecolor=blue,
    urlcolor=blue,
    citecolor=blue,
}
\usepackage{caption,setspace}
\usepackage{float}

\begin{document}
\begin{frontmatter}
\title{Multiple Sclerosis Lesion Synthesis in MRI using an encoder-decoder U-NET}
\author[label1,label2]{Mostafa Salem \corref{corr1}}
\author[label1]{Sergi Valverde}
\author[label1]{Mariano~Cabezas}
\author[label3]{Deborah Pareto}
\author[label1]{Arnau~Oliver}
\author[label1]{Joaquim Salvi}
\author[label3]{\`Alex~Rovira}
\author[label1]{Xavier~Llad{\'o}}

\address[label1]{Research Institute of Computer Vision and Robotics, University of Girona, Spain}
\address[label2]{Computer Science Department, Faculty of Computers and Information, Assiut University, Egypt}
\address[label3]{Magnetic Resonance Unit, Dept of Radiology, Vall d'Hebron University Hospital, Spain}

\cortext[corr1]{Corresponding author. M. Salem, Ed. P-IV, Campus Montilivi, University of Girona, 17003 Girona (Spain). e-mail: msalem@eia.udg.edu, mostafasalem@aun.edu.eg; Phone: +34 634878262; Fax: +34 972 418976.}

\begin{abstract}
In this paper, we propose generating synthetic multiple sclerosis (MS) lesions on MRI images with the final aim to improve the performance of supervised machine learning algorithms, therefore avoiding the problem of the lack of available ground truth. We propose a two-input two-output fully convolutional neural network model for MS lesion synthesis in MRI images. The lesion information is encoded as discrete binary intensity level masks passed to the model and stacked with the input images. The model is trained end-to-end without the need for manually annotating the lesions in the training set. We then perform the generation of synthetic lesions on healthy images via registration of patient images, which are subsequently used for data augmentation to increase the performance for supervised MS lesion detection algorithms. Our pipeline is evaluated on MS patient data from an in-house clinical dataset and the public ISBI2015 challenge dataset. The evaluation is based on measuring the similarities between the real and the synthetic images as well as in terms of lesion detection performance by segmenting both the original and synthetic images individually using a state-of-the-art segmentation framework. We also demonstrate the usage of synthetic MS lesions generated on healthy images as data augmentation. We analyze a scenario of limited training data (one-image training) to demonstrate the effect of the data augmentation on both datasets. Our results significantly show the effectiveness of the usage of synthetic MS lesion images. For the ISBI2015 challenge, our one-image model trained using only a single image plus the synthetic data augmentation strategy showed a performance similar to that of other CNN methods that were fully trained using the entire training set, yielding a comparable human expert rater performance.
\end{abstract}

\end{frontmatter}

\newpage
\section{Introduction}
\label{sec:introduction}
Multiple sclerosis (MS) is a disabling disease of the central nervous system that disrupts the flow of information within the brain and between the brain and body. It is characterized by the presence of lesions in the brain and spinal cord. Magnetic resonance imaging (MRI) has become one of the most important clinical tools to diagnose and monitor MS, since structural MRI depicts white matter (WM) lesions with high sensitivity \citep{Rovira2015}. The pattern and evolution of lesions has made MRI abnormalities invaluable criteria for the early diagnosis of MS. MRI allows high specificity and sensitivity visualization of the dissemination of WM lesions in time and space, which is a key factor in recent diagnostic criteria \citep{Filippi2016}. However, in both cross-sectional and longitudinal studies, manual or semiautomated segmentations have been used to compute the total number of lesions and the total lesion volume, which are challenging and time-consuming and prone to manual errors and inter- and intraobserver variability. This has lead to the development of different automated strategies \cite{LLADO2012164}.

Recently, deep neural networks have attracted substantial interest. Deep convolutional neural networks (CNN) have demonstrated groundbreaking performance in brain imaging, especially in tissue segmentation \citep{ZHANG2015214, Moeskops2016} and brain tumor segmentation \citep{KAMNITSAS201761,HAVAEI201718}. In contrast to previously supervised learning methods, CNNs do not require manual feature engineering or prior guidance. Furthermore, the increase in computing power makes them a very interesting alternative for automated lesion segmentation. CNN-based methods have achieved top ranking performance on all of the international MS lesion challenges \citep{Styner2008,commowick2016,ISBI2015,HashemiIEEEAccess}.

Studying MS lesions using supervised machine learning algorithms on MRI images requires a large number of samples to be annotated by expert radiologists. However, obtaining the annotations of medical images is time consuming. Several attempts have been made to overcome this challenge by using data augmentation. One of the most common data augmentation methods is to modify the dataset of images using geometric transformation such as image translation, rotation, or flip \citep{Krizhevsky2012}. However, the generated samples may not represent image appearances in real data, or the generated samples may be very similar to the existing images in the training dataset due to the parameters and image operators used \citep{Zhang_Chuanhai2017}. In contrast, we propose the generation of synthetic MS lesions on patient or healthy MRI images as the solution to the lack of expert annotations.

The synthesis of MRI images has attracted much interest in several areas of neuroimaging, including how to replace the missing MRI modalities with synthetic data \citep{Tulder2015}, to generate a subject-specific pathology-free image that is not present in the input modality \citep{Christopher2016}, to improve image segmentation and registration performance \citep{Iglesias2013} and others. The current state of the art in brain MRI synthesis is the work of \citet{MR_SYNTHESIS_TMI}. The authors proposed a deep fully convolutional neural network (FCNN) model for MRI synthesis, which takes different modalities as inputs and outputs synthetic images of the brain in one or more new modalities. This approach could be used for the synthesis of new lesions. However, there are some limitations that should be considered, such as the ability to control the intensity and the texture inside the lesions and the requirement of ground-truth masks for obtaining the lesion model.

In this paper, we propose a deep fully convolutional neural network model for MS lesion synthesis. The model takes as inputs T1-w and FLAIR images without MS lesions and outputs synthetic T1-w and FLAIR images with MS lesions. The MS lesion information is encoded as different binary masks passed to the model stacked with the input images. To overcome the limitations of the \citet{MR_SYNTHESIS_TMI} model, we divide the lesions into different regions based on voxel intensities, encoding this information as different binary masks. These binary masks are computed directly by thresholding the hyperintensities in the FLAIR image, so there is no need for the lesions' ground truth. That means the proposed MS lesion synthesis model is trained end-to-end without the need of manual expert MS lesion annotations in the training sets. Therefore, to tackle the lack of available ground-truth data needed for supervised MS lesion detection and segmentation strategies, we use the generated synthetic MS lesion images as data augmentation to improve the lesion detection and segmentation performance. This is done by synthesizing the lesions in new brain images, coming from either healthy subjects or from patients with lesions. Our evaluation included a clinical dataset and public MS data from the International Symposium on Biomedical Imaging (ISBI) 2015 MS challenge \citep{ISBI2015}. The accuracy of the generated synthetic images with MS lesions is evaluated qualitatively and quantitatively in terms of similarity performance and in terms of lesion detection and segmentation using a well-known state-of-the-art MS lesion segmentation method \citep{VALVERDE2017159}. For the data augmentation evaluation, we analyzed the effect of adding synthetic images on the segmentation performance while training with a different number of training images. To simulate a situation with very limited training data, we also analyzed the effect of the synthetic data augmentation starting from the one-image training scenario.

\section{Methods}
\label{materials_methods}
\subsection{MS lesion segmentation approach} 
The segmentation framework used for evaluating the proposed MS lesion generator is the state-of-the-art CNN model proposed by \citet{VALVERDE2017159}. Within this MS lesion segmentation framework, a cascade of two identical CNNs is optimized, where the first network is trained to be more sensitive to revealing possible candidate lesion voxels, while the second network is trained to reduce the number of false positive outcomes. For a complete description of the details and motivations for the proposed architecture, please refer to the original publication.

\subsection{Synthetic MS lesion generation pipeline}
\label{Synthetic_MS_lesions_pipeline}
\begin{figure*}[t]
    \centering
    \includegraphics[width=1.0\textwidth, keepaspectratio]{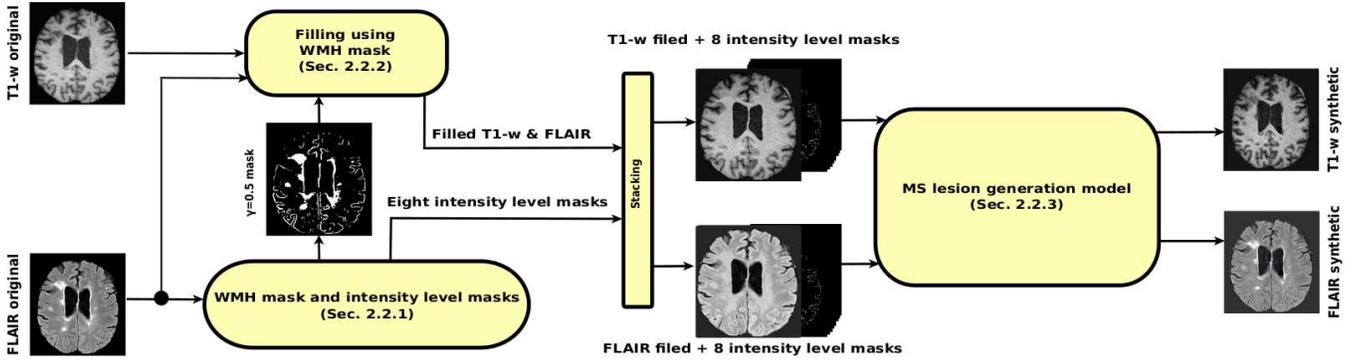}
    \caption{Scheme of the synthetic MS lesion generation pipeline. ${\gamma}=0.5$ WMH mask and the eight intensity level masks were computed by FLAIR thresholding. The ${\gamma}=0.5$ WMH mask was used to fill all the input modalities. Afterwards, the eight intensity level masks were stacked to each filled modality to create two 2D inputs with 9 channels each and these were the inputs to the MS lesions generator. For training, the original modalities were used as output. At testing time, if the intensity level masks were passed to the generator network without modification, the output images would be the generated version of the input ones containing all the WMHs found in the input image. Passing modified intensity level masks to the generator network will generate these modifications (i.e., new MS lesions) on the output images. }
    \label{generation_pipline}
\end{figure*}

To learn a model for the generation of synthetic MS lesions, images without lesions (used as inputs to the model) and the correspondent images with lesions (used as outputs to the model) are required. This kind of image set is not easy to obtain. One way to solve this would be using a longitudinal MS dataset; however, MS lesions in the baseline images and new MS lesions on the follow-up images should be annotated. Moreover, the baseline and follow-up images should also be registered. In that way, the model would be trained to generate new lesions in the follow-up scans. Nevertheless, in this scenario, new lesions on the follow-up images may not be sufficient to train the model since the volume of most of the new lesions can be relatively low \citep{SALEM2018}. Therefore, to overcome the lack of available ground-truth, we use the MS lesion generation pipeline shown in Figure \ref{generation_pipline} which consists of three main stages. First, the creation of an approximate white matter hyperintensity (WMH) mask and several intensity level masks to encode the intensity profile of the WMH voxels (Section~\ref{wm_hyperintense_masks}). Second, the filling of this WMH mask in the MR images with intensities resembling WM (Section~\ref{WMHs_filling}). Finally, the generation of MS lesions using the MS lesion generator network on the filled images (Section~\ref{lesions_generation_model}). Notice that the proposed MS generator was trained using only a cross-sectional MS dataset. These filled images were considered as images without lesions (used as inputs to the model), while the original images contained MS lesions (used as outputs to the model during the training process). The following subsections explain the full pipeline in more detail.
\subsubsection{WMH mask and intensity level masks}
\label{wm_hyperintense_masks} 
Creating the WMH mask and the intensity level masks is an important step in the proposed MS lesion generator pipeline. The aim is that training the model with intensity level masks instead of MS lesion masks avoids the limitation of having ground-truth. First, the FLAIR image is thresholded to obtain an approximate WMH mask. This mask is used to fill the WMH regions with intensities similar to the ones of the surrounding WM voxels. To learn the model for the generation of WMH voxels and their intensity profile, the range of intensities starting from the initial threshold is divided into different small ranges by increasing the intensity threshold at different steps. These created masks are considered as intensity level masks, which are then used to encode the intensity profile of the WMH voxels. The intensity level masks are stacked with the filled MR images when training the MS generator model. Therefore, the model can be trained with any dataset without requiring manual expert annotations.
The approximate WMH mask is computed by FLAIR thresholding. The threshold $T^F_{\gamma_{i}}$ and intensity level mask $IL_{i}$ are computed as follows:
\begin{center}
\begin{equation}
\label{WMHs_mask_computing}
T^F_{\gamma} = {\mu}^F_{GM} + \gamma {\sigma}^F_{GM}
\end{equation}
\end{center}
\begin{center}
\begin{equation}
\label{WMHs_mask_computing_2}
IL_{i} = T^F_{\gamma_{i}} < FLAIR \leq T^F_{\gamma_{i+1}}
\end{equation}
\end{center}
where ${\mu}^F_{GM}$ and ${\sigma}^F_{GM}$ are the intensity's distribution parameters of gray matter (GM) tissue on the FLAIR image \citep{CABEZAS2014147}. A small value of $\gamma$ must be chosen to obtain an approximate WMH mask so that all the WMH voxels are included in this mask. Different intensity level masks are obtained by increasing the $\gamma$ value. The higher the value of $\gamma$, the more brighter WMH voxels are included in the mask. 

\begin{figure*}[t]
  \begin{center}
    \includegraphics[width=1.0\textwidth]{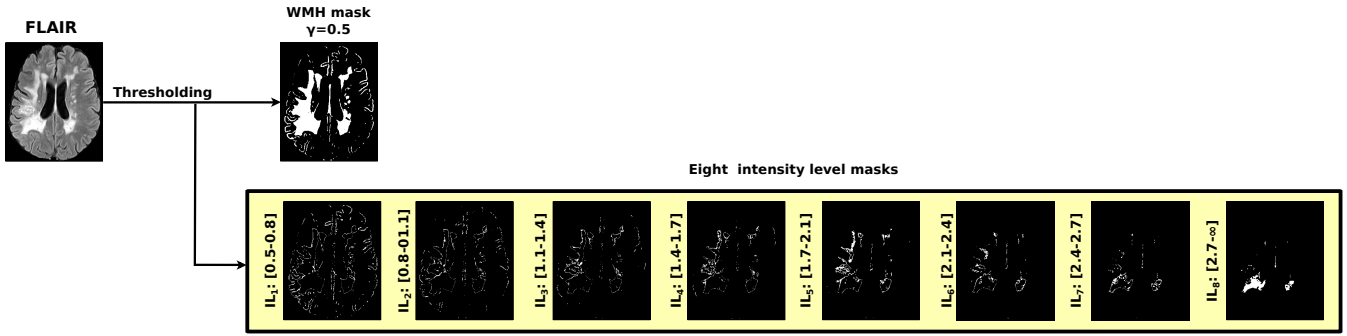}
  \end{center}
    \caption{The creation of the WMH mask and the eight intensity level masks (IL$_{1}$, IL$_{2}$, ..., and IL$_{8}$) using FLAIR thresholding.}
    \label{W1_8_generation}
\end{figure*}

In this study, the approximate WMH mask was obtained with $\gamma$ = 0.5. This value was found empirically to ensure that all the WMH voxels were included in the WMH mask. Eight intensity level masks with $\gamma=$ 0.5, 0.8, 1.1,  1.4, 1.7, 2.1, 2.4, and 2.7 were used to encode the WMH intensity profile. This was a trade-off between the memory required and the minimum number of training samples inside each intensity level mask while training the model. Figure \ref{W1_8_generation} describes the creation of the eight intensity level masks (IL$_{1}$, IL$_{2}$, ..., and IL$_{8}$). The ${ \gamma=0.5}$ WMH mask is used to fill the WMHs in the original image, and the intensity level masks are used to encode the intensity profile in the obtained WMH mask.
\subsubsection{WMH filling}
\label{WMHs_filling} 
After creating the intensity level masks described in the previous section, the ${ \gamma=0.5}$ WMH mask regions are filled in the input modalities. Similar to the work of \citet{FSL_Lesion_Filling}, a local filling method is used here to fill the WMH area with the surrounding WM voxels in all input modalities. First, for each slice in the MR image, the WMHs are split into individual connected regions. Second, each connected region is dilated twice. Each connected region is filled using values normally sampled using the mean and standard deviation of the WM voxels that were laid in the first dilated area. Furthermore, the filled area with its surrounding voxels (voxels in the filled connected region and the two dilated areas) is smoothed using a local Gaussian filter.

\subsubsection{MS lesion generation model}
\label{lesions_generation_model}
\begin{figure*}
    \centering
    \subfigure[The MS lesion generation model for two-input two-output case.]
    {
        \includegraphics[width=\textwidth, keepaspectratio]{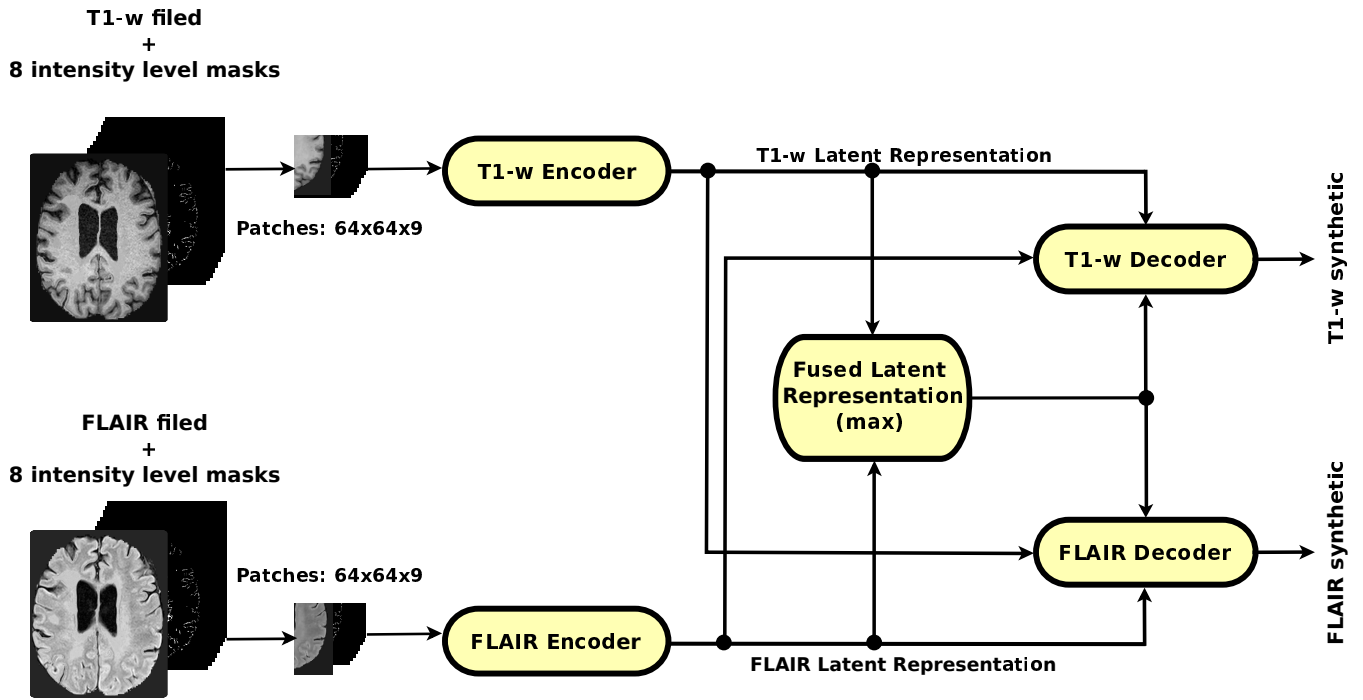}
        \label{encoder_decoder_a}
    }
    \\
    \subfigure[Encoder and decoder architectures.]
    {
        \includegraphics[width=\textwidth, keepaspectratio]{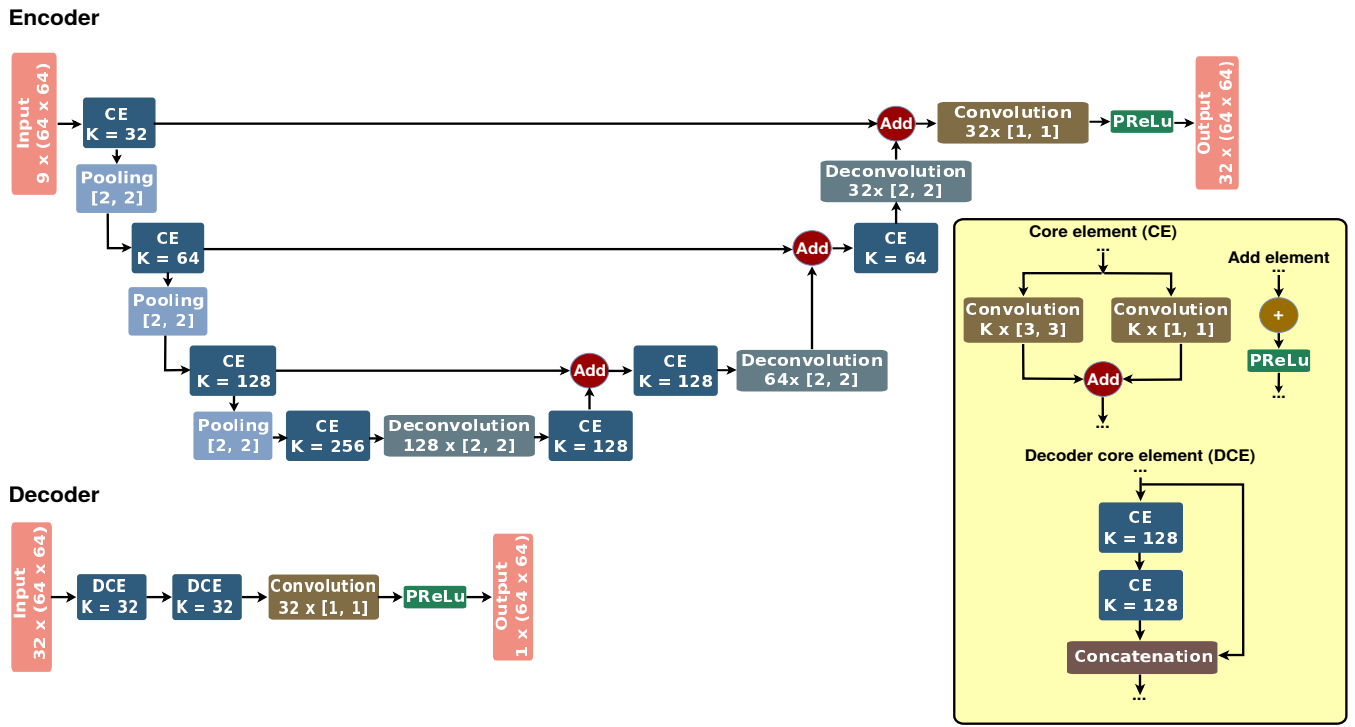}
        \label{encoder_decoder_b}
    }
\caption{MS lesion generator architecture. Each input modality has its own encoder that maps the input image modality to the 32-channel latent space. One decoder is learned for each output modality. The encoder maps the 32-channel latent representations to the outputs of that modality. Each decoder is used three times (i.e., shared decoder): once to decode each of the two individual latent representations (T1-w latent representation and FLAIR latent representation) and once to decode the fused representation. At testing time, we used the synthesis result from the fused representation as our output.}

    \label{encoder_decoder}
\end{figure*}

Figure \ref{encoder_decoder} shows our MS lesion generator architecture, which is inspired by the work of \citet{MR_SYNTHESIS_TMI}. As shown in Figure \ref{encoder_decoder_a}, it is a two-inputs-two-outputs model based on two encoders and two decoders (T1-w Encoder, FLAIR Encoder, T1-w Decoder, and FLAIR Decoder). The encoders are used to learn the latent representation for the input modalities, while the decoders are also used to generate the output modalities. Each decoder is used three times (i.e., shared decoder): one to decode each of the two individual latent representations (T1-w latent representation and FLAIR latent representation) and one to decode the fused latent representation. The fused latent representation is computed as the max function of the two individual latent representations. At testing time, we used the synthesis result from the fused latent representation as our output. The model has two 2D input patches with nine channels each (one input patch for each input modality). The eight intensity level masks computed as explained in Section~\ref{wm_hyperintense_masks} are stacked with each of the filled input modalities. The first channel is the filled image modality and the other eight channels are the intensity level masks. 

\textbf{Encoder architecture:} One independent encoder is built for each input modality following the architecture shown in Figure \ref{encoder_decoder_b}. The encoders embed input images into a latent space of 32-channel size. This architecture is inspired by the work of \citet{GUERRERO2018918}. It is a fully convolutional network that follows a U-shaped architecture \citep{U_net}. The U-Net's downsampling followed by the upsampling and skip connections allow the network to exploit information at large spatial scales, while not losing useful local information. Moreover, as discussed in \citet{drozdzal2016importance}, skip connections facilitate gradient flow during training. Our encoders are shallower than the original U-Net, having three downsample and upsample steps compared to the original four steps.

\textbf{Decoder architecture:} One decoder is built for each output modality following the architecture shown in Figure \ref{encoder_decoder_b}. The model is a fully convolutional network to map a multichannel image-sized latent representation to a single channel image of the required modality with synthetic MS lesions. 

\subsection{Data augmentation application: generating new synthetic MS lesions}
\label{MS_lesions_generation_on_new_dataset}

One of the applications of our synthetic MS lesion pipeline is to generate synthetic MS lesions on patient or healthy images and use these synthetic images as data augmentation to increase the MS lesion segmentation and detection performance. The main idea is to modify the original eight intensity level masks of the target image before passing it through the generator network. At testing time, if the intensity level masks are used without any modification, the output images are a generated synthetic version of the input ones containing all the WMHs found in the input image. Passing modified intensity level masks to the generator network will generate these desired modifications (i.e, new MS lesions) on the output images.

\begin{figure*}
  \begin{center}
    \includegraphics[width=\textwidth, height=\textheight,keepaspectratio]{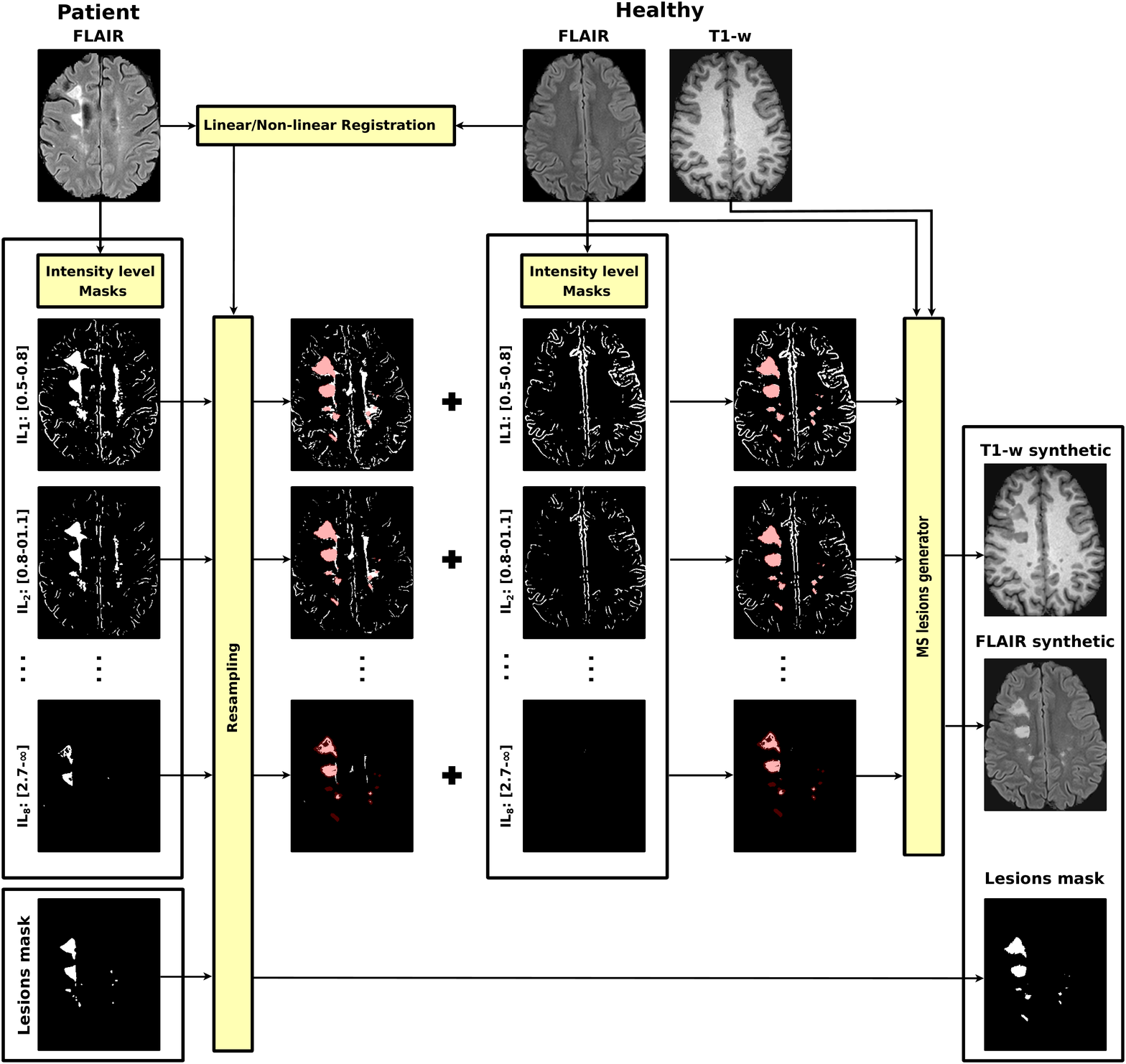}
  \end{center}
    \caption{Generating MS lesions on healthy subjects by linear/nonlinear registration. After registering the patient FLAIR to the healthy FLAIR, the lesion mask and the intensity level masks of the patient were resampled to the healthy space. The lesions from the patient resampled intensity level masks were copied to the healthy intensity level masks. The healthy images combined with their modified intensity level masks were passed to the MS lesions generated to generate the synthetic MS lesions on the healthy image.}
    \label{LesionsByRegistration}
\end{figure*}

Figure \ref{LesionsByRegistration} depicts how lesion expert annotations for a patient image can be generated on a healthy one through linear and nonlinear registration. After registration, the lesion mask and the eight intensity level masks of the patient subject are resampled to the healthy space. We split the resampled binary lesion mask into individual lesion volumes, in which every single lesion was defined as a spatially disconnected volume. After the lesion separation, the individual lesion volumes are dilated to incorporate the hyperintensities surrounding the lesions that are not annotated as lesion voxels. The intensity level masks of the dilated lesion volumes are copied from the patient resampled masks to the healthy masks. Finally, the healthy images plus their modified intensity level masks are passed through the generator network to add new MS lesions to the synthetic output images. In the same way, new MS lesions can be generated in patient images using patient-to-patient registration. Furthermore, more lesions could be added to the follow-up scans in the longitudinal MS analysis.
\section{Experimental setup}
\label{experimental_setup}

\subsection{Datasets}
\label{datasets_description}

\textbf{Clinical MS dataset:} This dataset consists of  15 healthy subjects and  65 different patients with a clinically isolated syndrome or early relapsing MS  (Vall d'Hebron Hospital Center, Barcelona, Spain) who underwent brain MR imaging for monitoring disease evolution and treatment response. Each patient underwent brain MRI within the first $3$ months after the onset of symptoms. The scans for all the patients were obtained in the same 3T magnet (Tim Trio; Siemens, Erlangen, Germany) with a 12-channel phased array head coil. The MRI protocol included the following sequences: 1) transverse proton density (PD)- and T2-w fast spin-echo (TR $=3080$~ms, TE $=21-91$~ms, voxel size $=0.78 \times 0.78 \times3.0$~mm$^3$), 2) transverse fast FLAIR (TR = $9000$~ms, TE $= 87$~ms, TI $= 2500$~ms, flip angle $= 120^{\circ}$, voxel size $= 0.49 \times 0.49  \times3.0$~mm$^3$), and 3) sagittal T1-w 3D magnetization-prepared rapid acquisition of gradient echo (TR $= 2300$~ms, TE $= 2.98$~ms, TI $= 900$~ms, voxel size $=1.0 \times 1.0 \times 1.2$~mm$^3$). The dataset was preprocessed as follows: for each patient, the T1-w image was linearly registered to the FLAIR using Nifty Reg tools\footnote{\url{https://sourceforge.net/projects/niftyreg/}} \citep{NiftyReg_1, NiftyReg_2}. Afterwards, a brain mask was identified and delineated on the registered T1-w image using the ROBEX Tool\footnote{\url{https://www.nitrc.org/projects/robex}} \citep{ROBEX}. Then, the two images underwent a bias field correction step using the N4 algorithm from the ITK library\footnote{\url{https://itk.org/Doxygen/html/classitk_1_1N4BiasFieldCorrectionImageFilter.html}} with the standard parameters for a maximum of 400 iterations \citep{Tustison2010}. 


\textbf{ISBI2015 dataset:} This dataset consists of 5 training and 14 testing subjects with 4 or 5 different image time-points per subject from the ISBI2015 MS lesion challenge \citep{ISBI2015}. Each scan was imaged and preprocessed in the same manner by the own organizers, with data acquired on a 3.0 Tesla MRI scanner (Philips Medical Systems, Best, The Netherlands) with T1-w MPRAGE, T2-w, PD and FLAIR sequences. For more information about the image protocol and preprocessing details, refer to the challenge organizers website\footnote{\url{http://iacl.ece.jhu.edu/index.php/MSChallenge/data}}. On the challenge competition, each subject image was evaluated independently, which led to a final training set and a testing set composed of 21 and 61 images, respectively. Additionally, manual delineations of MS lesions performed by two experts were included for each of the 21 training images.

For both datasets, brain tissue volume was computed using the FAST segmentation method \cite{FSL_FAST}. Finally, the WMH mask and the eight intensity level masks were computed by FLAIR thresholding as explained Section~\ref{wm_hyperintense_masks}, and T1-w and FLAIR were filled using the ${ \gamma=0.5}$ WMH mask computed using the method explained in Section~\ref{WMHs_filling}. 

\subsection{MS lesion generator training and implementation details}
\label{model_training}

To perform our experimental tests, we trained the lesion generator models into two different scenarios, one being the MS clinical dataset and the other one the ISBI2015 dataset (see Table \ref{datasets-table} for the images used for training). For training the generation network, 2D 64x64 patches with step size of 32x32 were extracted from the original images, the filled images, and the eight intensity level masks. The extracted patches were split into training and validation sets (70\% for training and 30\% for validation). The training set was used to adjust the weights of the neural network, while the validation set was used to measure how well the trained model was performing after each epoch. The extracted patches were passed to the network for training in mini batches of size 32 and the network was set to train for 200 epochs. To prevent overfitting, the training process was automatically terminated when the validation accuracy did not increase after 15 epochs. Regarding the MS lesion segmentation framework, the CNN training and inference procedures were identical to those proposed by \citet{VALVERDE2017159}.

The proposed method has been implemented in Python\footnote{\url{8 https://www.python.org}}, using Keras\footnote{\url{https://keras.io}} with the TensorFlow\footnote{\url{https://www.tensorflow.org/}} backend  \citep{tensorflow}. All experiments have been run on a GNU/Linux machine box running Ubuntu 18.04, with 128 GB RAM memory. The model training was carried out on a single TITAN-X GPU (NVIDIA Corp, United States) with 12 GB RAM memory. To promote the reproducibility and usability of our research, the proposed MS lesion generation pipeline is currently available for downloading at our research website\footnote{\url{https://github.com/NIC-VICOROB/MS_Lesions_Generator.}}.
\begin{table*}[t]
\scriptsize
\centering
\caption{Datasets. Total number of images, images used for training and testing the MS lesion generator, and images used for training and testing the MS lesion segmentation model for the clinical MS and ISBI2015 datasets.}
\label{datasets-table}
\begin{tabular}{c|l|l|l}
Datasets & Total number of images & MS lesion generator & MS lesion segmentation model\\ \Xhline{4\arrayrulewidth}
MS clinical dataset
&
\vtop
{\hbox {\strut - 65 patient images}
\hbox {\strut \hphantom{25} Group A (36 images)}
\hbox {\strut \hphantom{25} Group B (29 images)}
\hbox{\strut - 15 healthy images (VHhealthy)}
}   
&
\vtop{
\hbox {\strut }
\hbox {\strut Training: Group A (36 images)}
\hbox {\strut Testing: Group B (29 images)}
}  
&
\vtop{
\hbox {\strut Group B (29 images) is split into:}
\hbox {\strut \hphantom{25} Training: VHtrain (15 images)}
\hbox {\strut \hphantom{25} Testing: VHtest (14 images)}
} 
\\ \hline
ISBI2015 dataset
&
\vtop{
\hbox {\strut - 21 patient images (ISBItrain)}
\hbox {\strut - 61 patient images (ISBItest)}
}   
&
\vtop{
\hbox {\strut Training: ISBItest}
\hbox {\strut Testing: ISBItrain}
}  
&
\vtop{
\hbox {\strut Training: ISBItrain}
\hbox {\strut Testing: ISBItest}
} 
\\ \Xhline{4\arrayrulewidth}
\end{tabular}
\end{table*}
\subsection{Evaluation metrics}
\label{segmentation_framework_evaluation_metrics}

To evaluate the performance of the proposed MS lesion generator, we computed the similarity between the original and the synthetic images using the following similarity metrics:

\begin{itemize}
\item Mean Square Error (MSE):

\begin{equation*}
MSE (G, R) = \frac{1}{N} \sum_{i=1}^{N} (G_{i} - R_{i})
\end{equation*}

where $G$ and $R$ are the intensities of the generated and the real images, respectively, and $N$ is the number of voxels in the $R$ image.

\item Structural Similarity Index (SSIM):
\begin{equation*}
SSIM (G, R) = \frac{(2\mu_{G}\mu_{R} + c_{1}) (2\sigma_{GR} + c_{2})}{(\mu^{2}_{G} +\mu^{2}_{R} + c_{1})(\sigma^{2}_{G} +\sigma^{2}_{R} + c_{2})}
\end{equation*}
where ($\mu_{G}$, $\sigma^{2}_{G}$) and ($\mu_{R}$, $\sigma^{2}_{R}$) are the intensity's (mean, variance) of the generated and the real images, respectively, and $\sigma_{GR}$ is the covariance between them, $c_{1}$ and $c_{2}$ are two constants to stabilize the division with weak denominator.
\end{itemize}
On the other hand, the quantitative evaluation of the proposed MS lesion generator was performed by segmenting both the original and synthetic images individually using the same MS lesion segmentation framework and comparing the difference between the segmentation results. As explained before, the segmentation framework used to evaluate the proposed MS lesion generator is the MS lesion segmentation method proposed by \citet{VALVERDE2017159}, although the proposed data augmentation strategy could be applied to any approach. The evaluation of the resulting segmentations against the available lesion annotations was carried out using the following evaluation metrics:

\begin{itemize}
\item Dice Similarity Coefficient (DSC), which measures the overall segmentation accuracy between the manual lesion annotations and the output segmentation masks:

\begin{equation*}
DSC = \frac{2  \times TP_{s}}{2 \times TP_{s}+FP_{s}+FN_{s}}
\end{equation*}

where $TP_{s}$ and $FP_{s}$ denote the number of voxels correctly and incorrectly classified as a lesion, respectively, and $FN_{s}$ denotes the number of voxels incorrectly classified as a nonlesion.

\item Sensitivity of the method in detecting lesions between manual lesion annotations and output segmentation masks:
\begin{equation*}
sensitivity = \frac{TP_{d}}{TP_{d}+FN_{d}}
\end{equation*}

where $TP_{d}$ and $FN_{d}$ denote the number of correctly and missed lesion region candidates, respectively.

\item Precision of the method in detecting lesions between manual lesion annotations and output segmentation masks:
\begin{equation*}
precision = \frac{TP_{d}}{TP_{d}+FP_{d}}
\end{equation*}
where $TP_{d}$ and $FP_{d}$ denote the number of correctly and incorrectly classified lesion region candidates, respectively.
\end{itemize}
A paired t-test at the 5\% level was used to evaluate the significance of the data augmentation results. Significant results are shown in bold in all tables.

\section{Experiments and results}
\label{experiments_and_results}

\subsection{MS lesion synthesis}
\label{similarity_experiments}

In these experiments, qualitative and quantitative evaluations were undertaken by measuring the similarities between the real and the synthetic images in terms of MSE and SSIM metrics and in terms of MS lesion detection and segmentation using a state-of-the-art MS lesion segmentation method \citep{VALVERDE2017159} and the evaluation metrics described in section \ref{segmentation_framework_evaluation_metrics} (see Table \ref{datasets-table} for the images used).

\subsubsection{Evaluation}
\label{Evaluation1}

\textbf{Clinical MS dataset:} Both VHtrain and VHtest sets were generated using the proposed MS generator yielding VHtrainGen and VHtestGen, respectively. The evaluation of the proposed MS generator on this dataset was performed by measuring the MSE and SSIM metrics between the real and the synthetic images (using Group B images, see Table \ref{datasets-table}) and by training and testing the MS lesion segmentation model \citep{VALVERDE2017159} as follows: 1) training with the VHtrain set and testing on the VHtest set; 2) training with the VHtrainGen set and testing on the VHtestGen set; 3) training with the VHtrainGen set and testing on the VHtest set; and 4) training with the VHtrain set and testing on the VHtestGen set.

\textbf{ISBI2015 dataset:} The ISBItrain set was generated using the proposed MS generator yielding ISBItrainGen. Note that the evaluation of the ISBI 2015 challenge is performed blind by submitting the segmentation masks of the 61 testing cases to the challenge website evaluation platform\footnote{\url{https://smart-stats-tools.org/node/26}}. The evaluation of the proposed MS generator on this dataset was performed by measuring the MSE and SSIM metrics between the real and the synthetic images (using ISBItrain set, see Table \ref{datasets-table}). The performance of the two MS lesion segmentation models, one trained with the ISBItrain set and the other trained with the ISBItrainGen set, was evaluated by submitting to the challenge's evaluation platform, and comparing the accuracy between them.

\textbf{MS lesion generation on healthy subjects:} To evaluate the generation of MS lesions on healthy subjects by using registration, the MS lesions of the VHtrain dataset were generated on the VHhealthy images using linear and nonlinear registration as described in section \ref{MS_lesions_generation_on_new_dataset}. We refer to them as VHGenLinear and VHGenNonlinear, respectively. The evaluation of the proposed MS generator on these datasets was performed by training 3 MS lesion segmentation models using the VHGenLinear, the VHGenNonlinear, and (VHGenLinear + VHGenNonlinear) and testing on the VHtest set.

\subsubsection{Results}
\label{Results1}

\begin{figure*}
  \begin{center}
    \includegraphics[width=\textwidth, height=\textheight,keepaspectratio]{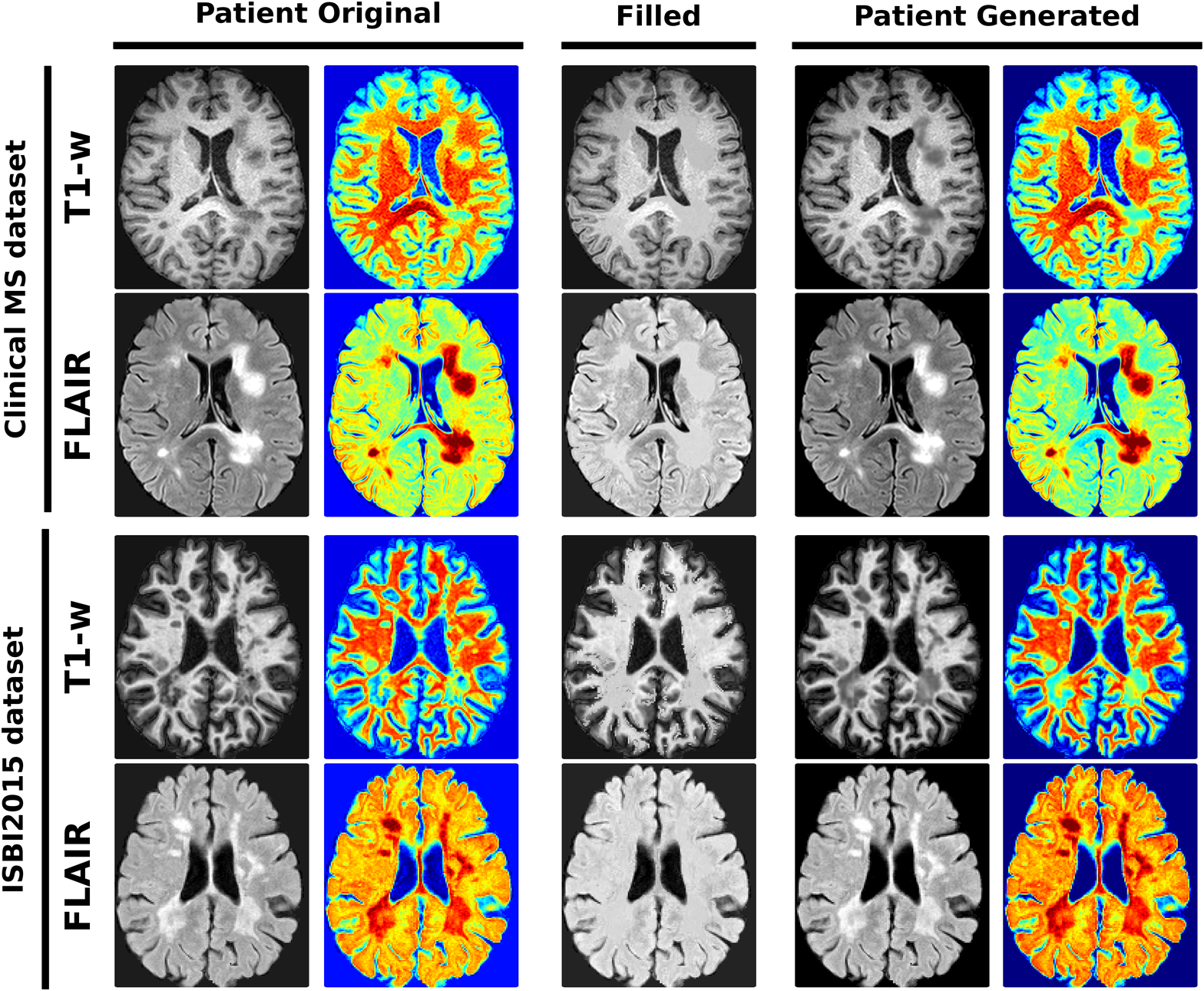}
  \end{center}
    \caption{Qualitative assessment of the proposed MS lesions generator. Slices are also displayed using jet color maps to visually enhance the intensities.}
    \label{LesionsGen_VH}
\end{figure*}

\begin{figure*}[t]
  \begin{center}
    \includegraphics[width=\textwidth,height=0.9\textheight,keepaspectratio]{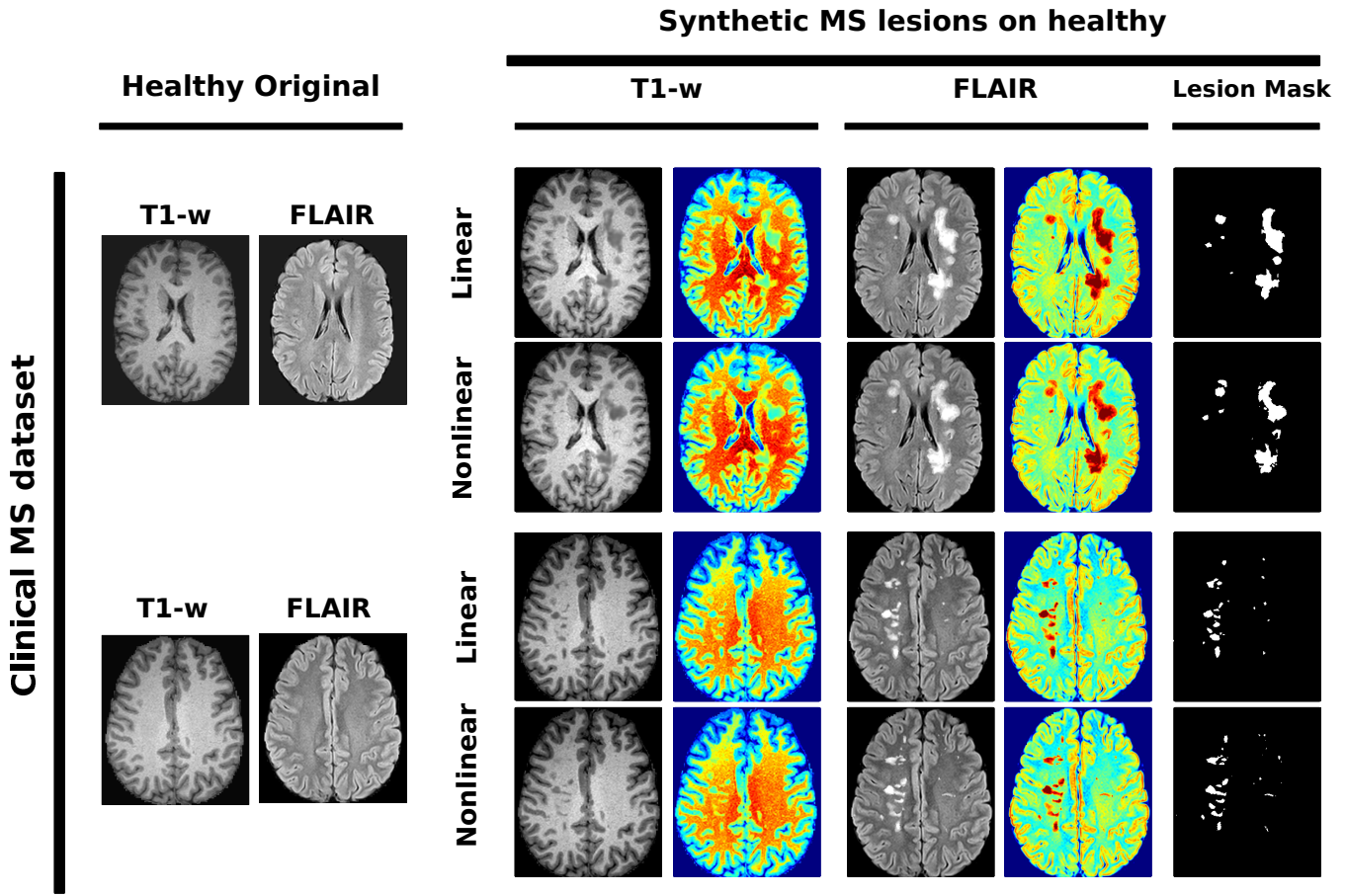}
  \end{center}
    \caption{Synthetic MS lesions generated on a healthy subject using linear/nonlinear registration. Slices are also displayed using jet color maps to visually enhance the intensities.}
    \label{LesionsGen_OnHealthy}
\end{figure*}

Table \ref{MSE-table} summarizes the MSE and SSIM between the real and synthetic images of the clinical MS and ISBI2015 datasets. Furthermore, the MSE and SSIM of ${ \gamma=0.5}$ WMH mask voxels are reported. Figure \ref{LesionsGen_VH} and \ref{LesionsGen_OnHealthy} show the qualitative assessment of the proposed MS lesion generator of the clinical MS/ISBI2015 datasets and synthetic MS lesions generated on healthy subjects using linear/nonlinear registration, respectively. The slices are also displayed using jet color maps to show the similarity of intensities inside the original and the synthetic lesions. Table \ref{comparison-table} summarizes the MS lesion detection and segmentation results, showing the obtained mean values when training with the original and synthetic images of the clinical MS and ISBI2015 datasets. The mean results when training with the synthetic MS lesions generated on healthy images using the clinical MS dataset lesion set are shown in Table \ref{comparisonHealthy-table}.
\begin{table}[t]
\scriptsize
\centering
\caption{Similarity results. MSE and SSIM between the original and synthetic images of the clinical MS (Group B set) and ISBI2015 (ISBItrain set) datasets for nonbackground and ${\gamma=0.5}$ WMH mask. The reported values are the mean $\pm$ standard deviation.}
\label{MSE-table}
\begin{tabular}{l|cc|cc}

\multicolumn{5}{c}{\textbf{Clinical MS Dataset (Group B set)}}                    \\ \Xhline{4\arrayrulewidth}
         & \multicolumn{2}{c|}{Non-background voxels} & \multicolumn{2}{c}{${ \gamma=0.5}$ WMH mask voxels} \\ \hline
				 & MSE                & SSIM              & MSE               & SSIM \\
T1-w       & $0.03 \pm\, 0.01$  & $0.96 \pm\, 0.01$ & $0.07 \pm\, 0.03$ & $0.93 \pm\, 0.03$  \\
FLAIR      & $0.02 \pm\, 0.01$  & $0.98 \pm\, 0.02$ & $0.03 \pm\, 0.07$ & $0.98 \pm\, 0.01$   \\

\Xhline{4\arrayrulewidth} \\

\multicolumn{5}{c}{\textbf{ISBI2015 Dataset (ISBItrain images)}}                    \\ \Xhline{4\arrayrulewidth}
         & \multicolumn{2}{c|}{Non-background voxels} & \multicolumn{2}{c}{${ \gamma=0.5}$ WMH mask voxels} \\ \hline
				 & MSE                & SSIM              & MSE               & SSIM     \\
T1-w      & $0.03 \pm\, 0.01$  & $0.97 \pm\, 0.01$ & $0.13 \pm\, 0.05$ & $0.94 \pm\, 0.03$   \\
FLAIR    & $0.01 \pm\, 0.01$   & $0.98 \pm\, 0.01$ & $0.01 \pm\, 0.01$ & $0.99 \pm\, 0.01$  \\

\Xhline{4\arrayrulewidth} \\
\end{tabular}
\end{table}
\begin{table}[t]
\scriptsize
\centering
\caption{Lesion segmentation and detection results. Comparison between the training using original images and synthetic images on Clinical MS and ISBI2015 datasets. For each coefficient ($DSC$, $sensitivity$, and $precision$), the reported values are the mean $\pm$ standard deviation. For the ISBI2015 dataset, the reported values are extracted from the challenge results board.}
\label{comparison-table}
\begin{tabular}{lcccc}
\multicolumn{4}{c}{\textbf{Clinical MS Dataset}}                                     \\ \Xhline{4\arrayrulewidth}
Train/Test        & $DSC$                & $sensitivity$       & $precision$                  \\\hline

VHtrain/VHtest         & $0.70 \pm\, 0.16$    & $0.69 \pm\, 0.13$   & $0.73 \pm\, 0.15$      \\ 

VHtrainGen/VHtest      & $ 0.68 \pm 0.16 $    & $ 0.72 \pm 0.14 $   & $ 0.71 \pm 0.13$      \\
VHtrainGen/VHtestGen   & $0.67 \pm\, 0.17$    & $0.65 \pm\, 0.14$   & $0.70 \pm\, 0.17$        \\  
VHtrain/VHtestGen   & $0.68 \pm\, 0.15$    & $0.66 \pm\, 0.15$   & $0.70 \pm\, 0.16$        \\  
\hline

\Xhline{4\arrayrulewidth} \\

\multicolumn{4}{c}{\textbf{ISBI2015 Dataset}}                    \\ \Xhline{4\arrayrulewidth}

Train/Test       & $DSC$                & $sensitivity$        & $precision$                   \\ \hline
ISBItrain/ISBItest    & $ 0.64 \pm 0.12 $    & $ 0.57 \pm 0.16 $    & $ 0.79 \pm 0.12$          \\
ISBItrainGen/ISBItest & $ 0.64 \pm 0.13 $    & $ 0.56 \pm 0.17 $    & $ 0.80 \pm 0.14$        
\\ \Xhline{4\arrayrulewidth} \\

\end{tabular}
\end{table}
\begin{table}[t]
\scriptsize
\centering
\caption{Clinical MS dataset results of training using synthetic images generated on healthy subjects as described in section \ref{MS_lesions_generation_on_new_dataset}. For each coefficient ($DSC$, $sensitivity$, and $precision$), the reported values are the mean $\pm$ standard deviation.}
\label{comparisonHealthy-table}
\begin{tabular}{lccc}

\Xhline{4\arrayrulewidth}
Train/Test             & $DSC$                & $sensitivity$         & $precision$                  \\\hline
VHGenLinear/VHtest                 & $ 0.63 \pm 0.21 $    & $ 0.63 \pm 0.17 $     & $ 0.63 \pm 0.16 $      \\
VHGenNonlinear/VHtest              & $ 0.63 \pm 0.20 $    & $ 0.62 \pm 0.14 $     & $ 0.62 \pm 0.16 $        \\
Both/VHtest                      & $ 0.65 \pm 0.20 $    & $ 0.64 \pm 0.14 $     &  $ 0.64 \pm 0.17 $       \\
\Xhline{4\arrayrulewidth}
\end{tabular}
\end{table}
\subsection{Data augmentation experiments}
\label{data_augmentation_experiments}
In these experiments, we evaluated the use of the proposed MS lesion generator as a data augmentation method by generating the lesion masks on healthy images from the same domain using registration as described in section \ref{MS_lesions_generation_on_new_dataset}. The two deformed generated lesion masks (from linear and nonlinear registration) and the correspondent two synthetic images were added to the original patient image during training as data augmentation.
\subsubsection{Evaluation}
\label{Evaluation2}
\textbf{Clinical MS dataset:} For each patient image from the VHtrain set, we created two synthetic images with lesions on a healthy image from the VHhealthy set (VHGenLinear and VHGenNonlinear) as described in section~\ref{MS_lesions_generation_on_new_dataset}. Those two synthetic images were used together with the original image as data augmentation in the following experimental tests: 1) to analyze the effect of the synthetic data augmentation images on the segmentation performance while training with different number of training images, two models were trained using 1, 2, 3, 5, 10 or all of the available training images, with one model using the original images and the other using the same original images plus their synthetic data augmentation images; and 2) to simulate a situation with limited training data, we analyzed the effect of the synthetic data augmentation on the segmentation performance in the scenario of having only one-image for training. Using a single training image with a lesion volume in the range of $0.34-49.4$ ml, two models were trained. One model used the original image (i.e., from VHtrain) and the other used the same original image plus the two synthetic images generated on the healthy image (i.e., from VHGenLinear and VHGenNonlinear).

\textbf{ISBI2015 dataset:} To simulate a situation with limited training data, we analyzed the effect of the synthetic data augmentation images on the segmentation performance in the one-image training scenario on the overall performance of the testing set. To do so, we chose a single training image from each training subject (ISBItrain), which led to 5 different training sets with a varying number of lesions and a total lesion volume in the range $2.3-26.8$ ml. Since there were no healthy subjects available from this challenge, we chose the fourth training subject (this image has the smallest lesion load; $\approx 2.3$ ml) and filled it as described in section \ref{wm_hyperintense_masks} (but only MS lesions were filled instead of the WMH areas). We considered this image as a healthy subject and we refer to it as ISBI-H. The MS lesions of each of the four selected ISBI images were generated on the ISBI-H using linear and nonlinear registration, as described in section \ref{MS_lesions_generation_on_new_dataset}, yielding, for each patient image from the selected four, two generated images and their correspondent lesion masks that were used as data augmentation. Based on this, we undertook the following experiments. 1) To simulate a situation with limited training data, we analyzed the effect of the synthetic data augmentation images on the segmentation performance in the one-image training scenario. Using a single training image from the four images selected, two models were trained, one using the original image and the other using the original image plus its two synthetic images generated on ISBI-H using linear and nonlinear registration. 2) To determine the performance of all the models trained on the blind test set, all trained models from the previous experiment were sent to the challenge's evaluation platform, comparing its accuracy to those of the other submitted MS lesion segmentation pipelines fully trained using the entire available training set. Among the set of evaluated coefficients computed in the challenge, only the DSC, sensitivity and precision metrics are shown for comparison.

\subsubsection{Results}
\label{Results2}
\begin{figure*}
    \centering
    \subfigure[]
    {
        \includegraphics[width=.30\textwidth]{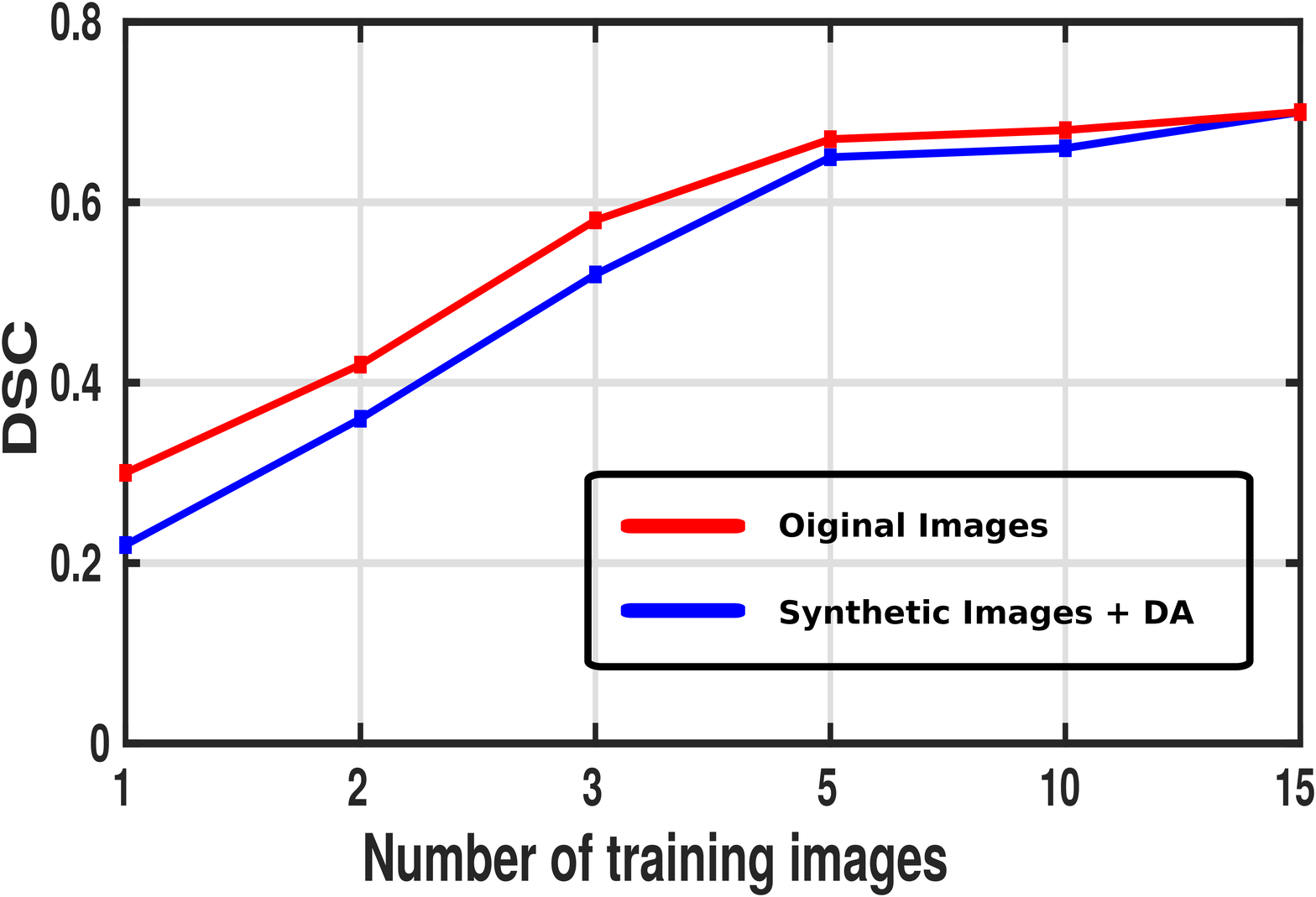}
        \label{VH_Augmentation_a}
    }    
    \subfigure[]
    {
        \includegraphics[width=.30\textwidth]{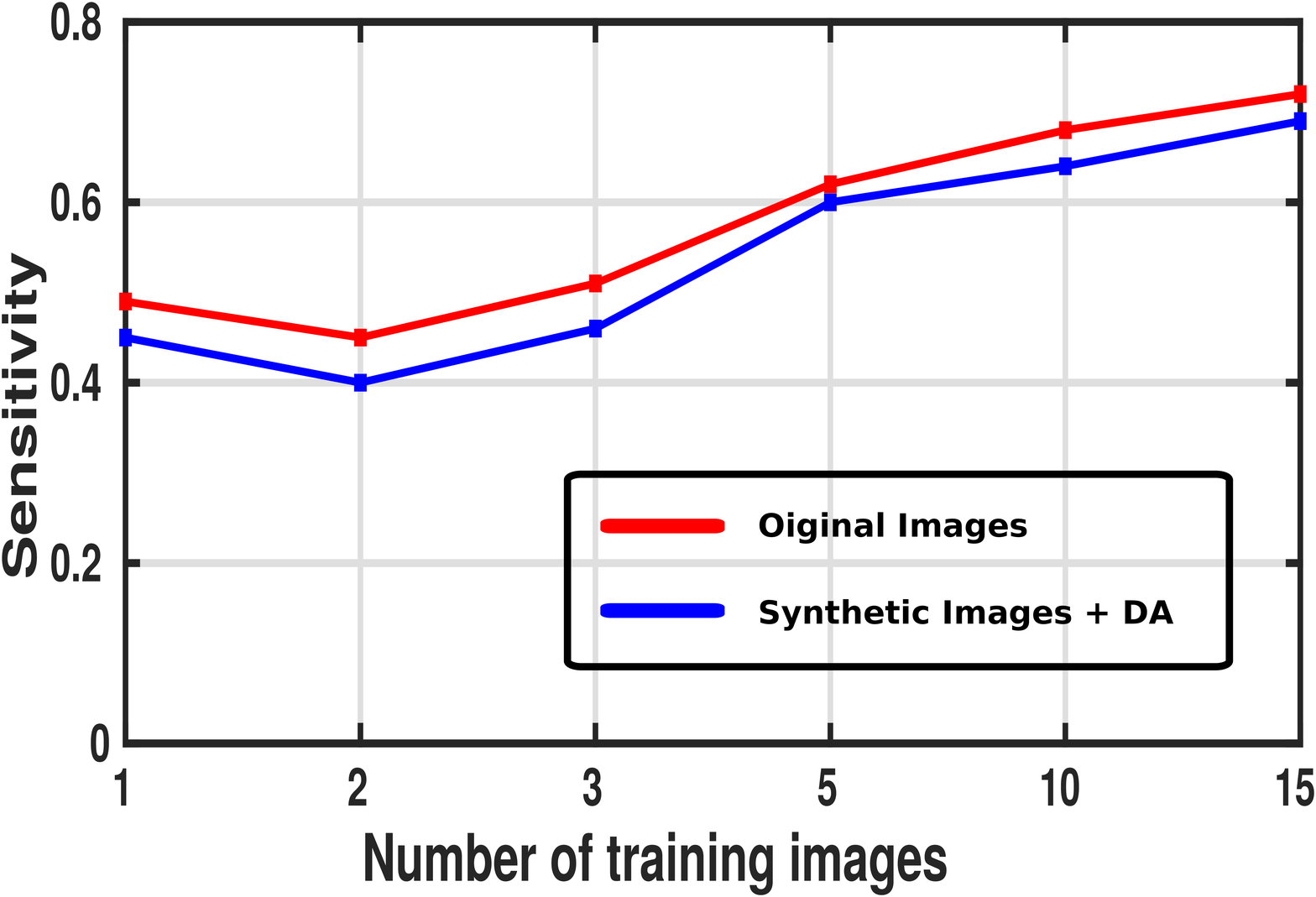}
        \label{VH_Augmentation_b}
    }    
    \subfigure[]
    {
        \includegraphics[width=.30\textwidth]{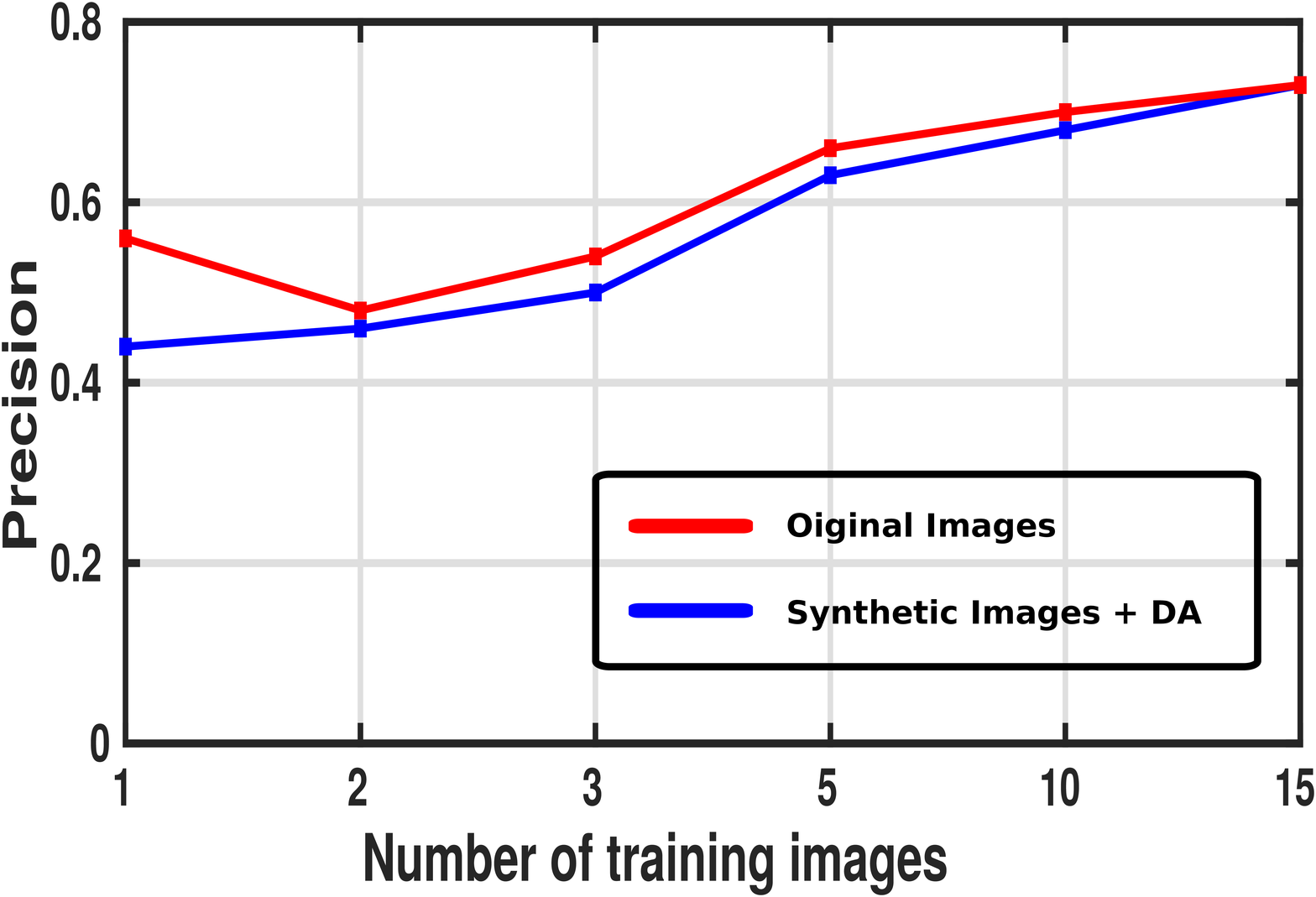}
        \label{VH_Augmentation_c}
    }    
    \caption{Effect of the number of training images and their DA images on the DSC, sensitivity and precision coefficients when evaluated on the clinical MS dataset. The represented value for each configuration is computed as the mean DSC, sensitivity and precision scores over the 14 VHtest images.}
    \label{VH_Augmentation}
\end{figure*}

Regarding the Clinical MS dataset, Figure \ref{VH_Augmentation} shows the DSC, sensitivity and precision coefficients of different models trained using different number of training images, which ranged from 1 to 15 images. Table \ref{VH-OneShot-table} shows the DSC, sensitivity and precision coefficients of the models under the one-image training scenario. Regarding the ISBI2015 dataset, Table \ref{ISBI-OneShot-blindevaluatoin} shows the performance of each of the one-image scenario models when trained on different images with varying degrees of lesion size. Table \ref{ISBI-OneShot-comparison} shows the performance of the models trained with ISBI02 plus DA against different top rank participant challenge strategies. From the list of compared methods, the best five strategies were based on CNN models (\citet{Andermatt2017,Salehi2017,VALVERDE2017159,Birenbaum2017}), while the others were based on either other supervised learning techniques (\citet{Valcarcel2018,DESHPANDE20152,Sudre2015}) or unsupervised intensity models (\citet{SHIEE20101524,JAIN2015367}). 

\begin{table*}[t]
\scriptsize
\centering
\caption{One-image scenario for the Clinical MS dataset: $DSC$, $sensitivity$ and $precision$ coefficients for two models, one model trained using a single original image (ORG) and the other one trained using same single image plus its synthetic data augmentation images (DA) with varying degrees of lesion load. For each coefficient, the reported values are the mean $\pm$ standard deviation when evaluated on the VHtest set.}
\label{VH-OneShot-table}
\begin{tabular}{llccc}
lesion vol (num lesions)   & & $DSC$              & $sensitivity$         & $precision$            \\\hline
\Xhline{4\arrayrulewidth}

0.34 ml (18 lesions) & ORG & $ 0.18 \pm 0.11 $     & $ 0.43 \pm 0.13 $      & $ 0.41 \pm 0.11 $     \\
                     & DA  & $ \mathbf{0.29 \pm 0.14} $     & $ \mathbf{0.48 \pm 0.15}$  & $ \mathbf{0.61 \pm 0.20} $      \\\hline

1.0 ml (6 lesions)  & ORG & $ 0.35 \pm 0.25 $     & $ 0.23 \pm 0.15 $     & $ 0.35 \pm 0.29 $      \\
					& DA  & $ \mathbf{0.47 \pm 0.25} $     & $ \mathbf{0.27 \pm 0.14} $     & $ 0.37 \pm 0.21 $      \\\hline

2.0 ml (25 lesions) & ORG & $ 0.53 \pm 0.19 $     & $ 0.43 \pm 0.16 $     & $ 0.62 \pm 0.26 $      \\
					& DA  & $ \mathbf{0.57 \pm 0.20} $     & $ \mathbf{0.54 \pm 0.14} $     & $ 0.64 \pm 0.28 $      \\\hline

5.5 ml (15 lesions)  & ORG & $ 0.28 \pm 0.15 $     & $ 0.28 \pm 0.12 $     & $ 0.32 \pm 0.14 $     \\
					 & DA  & $ \mathbf{0.32 \pm 0.12} $     & $ \mathbf{0.35 \pm 0.13} $     & $ \mathbf{0.38 \pm 0.12} $      \\\hline

7.6 ml (42 lesions)  & ORG & $ 0.57 \pm 0.25 $     & $ 0.41 \pm 0.16 $     & $ 0.53 \pm 0.23 $     \\
					 & DA  & $ \mathbf{0.63 \pm 0.20} $     & $ \mathbf{0.50 \pm 0.16} $      & $ \mathbf{0.65 \pm 0.21} $      \\\hline

21.5 ml (181 lesions) & ORG & $ 0.61 \pm 0.21 $     & $ 0.61 \pm 0.18 $     & $ 0.54 \pm 0.14 $     \\
					  & DA  & $ 0.63 \pm 0.20 $     & $ \mathbf{0.66 \pm 0.14} $     & $ \mathbf{0.60 \pm 0.14} $      \\\hline

49.4 ml (53 lesions) & ORG & $ 0.57 \pm 0.25 $     & $ 0.58 \pm 0.20 $     & $ 0.56 \pm 0.22 $     \\
					 & DA  & $ 0.58 \pm 0.25 $     & $ \mathbf{0.67 \pm 0.18} $     & $ \mathbf{0.60 \pm 0.13} $      \\
\Xhline{4\arrayrulewidth}
\end{tabular}
\end{table*}
\begin{table*}[t]
\scriptsize
\centering
\caption{One-image scenario for the ISBI2015 dataset: $DSC$, $sensitivity$, $precision$, and overall score coefficients for two models, one model trained using a single original image (ORG) and the other one trained using same single image plus its synthetic data augmentation images (DA). The reported values are extracted from the challenge results board. For each coefficient, the reported values are the mean $\pm$ standard deviation when evaluated on the ISBItest set.}
\label{ISBI-OneShot-blindevaluatoin}
\begin{tabular}{llcccc}
lesion vol (num lesions)     &      & $DSC$  & $sensitivity$          & $precision$     & $score$ \\
\Xhline{4\arrayrulewidth}

ISBI01 & ORG         & $ 0.41 \pm 0.13 $      & $ 0.30 \pm 0.12 $      & $ 0.75 \pm 0.19 $  & $87.60$ \\    
       & DA    & $ \mathbf{0.54 \pm 0.13 }$    & $ \mathbf{0.45 \pm 0.15 }$       & $ 0.75 \pm 0.17 $  & $89.54$  \\   \hline

ISBI02 & ORG         & $ 0.53 \pm 0.18 $      & $ 0.44 \pm 0.19 $      & $ 0.76 \pm 0.21 $  &$88.60$      \\
       & DA    & $ \mathbf{0.59 \pm 0.15} $     & $ \mathbf{0.51 \pm 0.19} $     & $ \mathbf{0.78 \pm 0.18} $   & $90.05$ \\  \hline

ISBI03 & ORG  & $ 0.49 \pm 0.13 $     & $ 0.39 \pm 0.14 $  & $ 0.74 \pm 0.18 $    &  $88.67$ \\
       & DA   & $ 0.49 \pm 0.12 $   & $ 0.39 \pm 0.14 $     & $ 0.\mathbf{77 \pm 0.15} $ & $89.55$\\    
                             \hline

ISBI05  & ORG  & $ 0.39 \pm 0.13 $  & $ 0.29 \pm 0.13 $      & $ 0.73 \pm 0.17 $ &$88.02$      \\
        & DA   & $ \mathbf{0.42 \pm 0.13} $  & $ 0.30 \pm 0.12 $     & $ \mathbf{0.79 \pm 0.16} $ & $88.66$\\ 
\Xhline{4\arrayrulewidth}
\end{tabular}
\end{table*}
\begin{table*}[t]
\scriptsize
\centering
\caption{ISBI2015 challenge: $DSC$, $sensitivity$, $precision$ and overall score coefficients for the best one-image scenario with the data augmentation model (ISBI02 $+$ DA). The obtained results are compared with different top rank participant strategies. For each method, the reported values are extracted from the challenge results board. The reported values are the mean (standard deviation) when evaluated on the 61 testing images. The performance of the methods with an overall $score \geq 90$ is considered to be similar to human performance.}
\label{ISBI-OneShot-comparison}
\begin{tabular}{lcccc}
Method      &   $DSC$     & $sensitivity$         & $precision$  & $score$                \\ \Xhline{4\arrayrulewidth}
\citet{Andermatt2017}    & $ 0.63 \pm 0.14 $ & $ 0.54 \pm 0.19 $ & $ 0.84 \pm 0.10 $&  $ 92.07$ \\
\citet{Salehi2017}       & $ 0.66 \pm 0.11 $ & $ 0.67 \pm 0.20 $ & $ 0.71 \pm 0.16 $&  $ 91.52$ \\
\citet{VALVERDE2017159}  & $ 0.64 \pm 0.12 $ & $ 0.57 \pm 0.17 $ & $ 0.79 \pm 0.15 $&  $ 91.44$ \\
\citet{Birenbaum2017}    & $ 0.63 \pm 0.14 $ & $ 0.55 \pm 0.18 $ & $ 0.80 \pm 0.15 $&  $ 91.26$ \\
\citet{DESHPANDE20152}   & $ 0.60 \pm 0.13 $ & $ 0.55 \pm 0.17 $ & $ 0.73 \pm 0.18 $&  $ 89.81$ \\
\citet{JAIN2015367}      & $ 0.55 \pm 0.14 $ & $ 0.47 \pm 0.15 $ & $ 0.73 \pm 0.20 $&  $ 88.74$ \\
\citet{SHIEE20101524}    & $ 0.55 \pm 0.19 $ & $ 0.54 \pm 0.15 $ & $ 0.70 \pm 0.29 $&  $ 88.46$ \\
\citet{Valcarcel2018}    & $ 0.57 \pm 0.13 $ & $ 0.57 \pm 0.18 $ & $ 0.61 \pm 0.16 $&  $ 87.71$ \\
\citet{Sudre2015}        & $ 0.52 \pm 0.14 $ & $ 0.46 \pm 0.15 $ & $ 0.66 \pm 0.18 $&  $ 86.44$ 
 \\ \Xhline{4\arrayrulewidth}
ISBI02 + DA              & $ 0.59 \pm 0.15 $ & $ 0.51 \pm 0.19 $ & $ 0.78 \pm 0.18 $&  $90.05$ \\
\Xhline{4\arrayrulewidth}
\end{tabular}
\end{table*}
\section{Discussion and future work}
We proposed a synthetic MS lesion generator pipeline that generates synthetic images with MS lesions. The use of the intensity level masks introduced in our proposal enabled us to train the model without the need of ground truth. Furthermore, the intensity level masks help the MS lesion generator to preserve the intensity gradients inside the synthetic MS lesion. Although the proposed pipeline was used to generate MS lesions on T1-w and FLAIR images using only two encoders and two decoders, the model can be easily extended to new input/output modalities through the addition of new encoders/decoders.

We demonstrated the similarity between the synthetic and real lesions qualitatively and quantitatively on patient and healthy subjects. Synthetic images are very similar to the real ones in terms of the two similarity metrics for nonbackground and ${\gamma=0.5}$ WMH mask voxels for both datasets. Regarding the MS lesion segmentation results, the experiments show how similar the training is using real or synthetic images in terms of MS lesion detection. Regarding the MS clinical dataset, the performance is 2\% less in terms of DSC and precision when training with the synthetic images than training with the real images. However, similar results were obtained when training with real images and testing on synthetic images. From the results obtained, synthetic images could be used as training or testing images. Regarding the ISBI2015 datasets, the performance is very similar in terms of the three coefficients. Regarding the training using synthetic MS lesions generated on healthy subjects, good segmentation and detection results were obtained when training with synthetic images generated on healthy subjects. The performance is also  very similar when training with synthetic images generated using linear, nonlinear registration or both. 

Regarding the data augmentation experiments, we demonstrated the effect of data augmentation on the MS lesion segmentation performance when increasing the number of the training images. The difference in performance between training with original images and original images plus DA decreases in terms of the three metric coefficients as the number of the training images increases. The DA images generated from linear and nonlinear registration do not give more variability to the training data when increasing the number of training images. Furthermore, to simulate a situation with limited training data, we analyzed the effect of one-image training scenario. Regarding the MS clinical dataset, significant improvement was obtained in terms of the three metric coefficients with a lesion volume in the range of $0.34-49.4$ ml. Regarding the ISBI2015 dataset, a significant improvement was obtained in terms of the three metric coefficients, except for ISBI03, where only a significant improvement in precision was obtained. Comparing the accuracy of the best performing model (ISBI02+DA) to those of the other submitted MS lesion segmentation pipelines fully trained using the entire available training set, the proposed one image plus its data augmentation images reported a performance similar to that of the same fully trained cascaded CNN architecture (score $91.44$) \citep{VALVERDE2017159}, which shows the improvement of the proposed data augmentation strategy to the training used with limited training data.

Currently, work is underway to build a lesion dictionary containing the MS lesion information (the annotation and the intensity level masks) of different MS lesions grouped by lesions load, and the extension could be an automatic selection of suitable insertion location so that the lesions selected from the dictionary could be generated synthetically in multiple locations without manual user involvement. Choosing the automatic locations of lesions is not an easy task because inserting lesions in incorrect locations may mislead the training process and decrease the overall performance. We believe that generating synthetic MS lesions on healthy subjects using the dictionary and the automatic locations will provide more variability to the training data than the linear/nonlinear registration, as data augmentation and the overall performance of the proposed pipeline will improve accordingly.

In conclusion, the obtained results indicate that the proposed pipeline is able to generate useful T1-w and FLAIR synthetic images with MS lesions that do not differ from real images. Furthermore, the combination of the synthetic MS lesions generated on healthy images and original patient images from the same domain increases the segmentation and detection accuracy of MS lesions.

\section*{Acknowledgment}
Mostafa Salem holds a grant for obtaining the Ph.D. degree from the Egyptian Ministry of Higher Education. This work has been partially supported by La Fundaci\'{o} la Marat\'{o} de TV3, by Retos de Investigaci\'{o}n TIN2014- 55710-R, TIN2015-73563-JIN and DPI2017-86696-R from the Ministerio de Ciencia y Tecnolog\'{i}a. The authors gratefully acknowledge the support of NVIDIA Corporation with their donation of the TITAN-X PASCAL GPU used in this research. 

\section*{References}
\bibliographystyle{abbrvnat}

\end{document}